\documentclass[a4paper,11pt]{article}
\usepackage[a4paper,margin=1in]{geometry}
\usepackage{graphicx}
\usepackage{inconsolata}
\usepackage{mathpazo}
\usepackage{booktabs}
\usepackage{amsmath}
\usepackage[sort]{natbib}
\usepackage{hyperref}
\usepackage[dvipsnames]{xcolor}
\usepackage{multirow}
\usepackage{makecell}
\usepackage{titlesec}
\usepackage{lipsum}
\usepackage{etoolbox}
\usepackage{hyperref}
\usepackage{setspace}
\usepackage{lineno}
\usepackage{fancyhdr}
\usepackage{enumitem}
\usepackage[normalem]{ulem}
\setlist{label*=(\arabic*)}
\usepackage[linguistics]{forest}

\makeatletter
\patchcmd{\@maketitle}{\LARGE}{\Large}{}{}
\makeatother

\titleformat{\section}
  {\normalfont\large\bfseries}{\thesection}{1em}{}

\titleformat{\subsection}
  {\normalfont\normalsize\bfseries}{\thesubsection}{1em}{}

\titleformat{\subsubsection}
  {\normalfont\normalsize\bfseries}{\thesubsubsection}{1em}{}

\newcommand{\mplus}{\text{ + }}
\newcommand{\bdot}{\boldsymbol{\cdot}}

\title{}
\title{Why are language models less surprised than humans? Testing the Parse Multiplicity Mismatch Hypothesis} 
\author{William Timkey$^{1}$\footnotemark \enspace, Brian Dillon$^{3}$ \enspace, Tal Linzen$^{1,2}$}

\footnotetext[1]{Department of Linguistics, New York University, 10 Washington Pl, New York, NY 10003, USA}
\footnotetext[2]{Center for Data Science, New York University, 60 5th Ave, New York, NY 10011, USA}
\footnotetext[3]{Department of Linguistics, University of Massachusetts Amherst, 650 N Pleasant St, Amherst, MA 01002, USA}

\footnotetext[1]{Correspondence: wpt2011@nyu.edu}

\date{May 2026}

\begin{document}
\pagestyle{fancy}
\fancyhead{} %
\setlength{\footskip}{14pt}

\maketitle
\section*{Abstract}
Surprisal theory posits that the processing difficulty of a word is determined by its predictability in context, offering a potential link between human sentence processing and next-word predictions from language models. While language model (LM) surprisals successfully predict reading times in naturalistic text, they systematically underpredict the magnitude of difficulty observed in controlled studies of syntactic ambiguity, particularly in garden path sentences. This mismatch might arise from differences in the computational constraints between humans and LMs. Here we test one such hypothesis, specifically, that LMs may be able to simultaneously consider a greater number of distinct sentence interpretations at once, compared to humans. Using Recurrent Neural Network Grammars (RNNGs) with word-synchronous beam search, we systematically vary the number of simultaneous parses used to compute word surprisal, and then use these surprisals to predict human reading times. Reducing the number of simultaneous active parses indeed increases the magnitude of predicted garden path effects, but not nearly enough to capture the full magnitude of the effects in humans. This suggests that differences in the number of simultaneous parses available to LMs and humans cannot reconcile LM-based surprisal with human sentence processing.

\newpage

\section*{Acknowledgments}
Many thanks to Qingyang Zhu and members of the NYU Computation and Psycholinguistics lab for their insightful feedback and discussion. This project is supported by the National Science Foundation (NSF) under grants BCS-2020914, BCS-2020945, IIS-2504953 and IIS-2504954. This work was supported in part through the NYU IT High Performance Computing resources, services, and staff expertise. B.D. is additionally supported by the Samuel F. Conti Fellowship from the University of Massachusetts, Amherst.

\doublespacing
\newpage

\section{Introduction}
Prediction is widely regarded to be a central organizing principle of human cognition in general and language in particular. Suggestive of this possibility are the recent successes of large language models, which have demonstrated remarkable language understanding capabilities via the simple objective of predicting the next word in a sentence. The advent of such models has led many psycholinguists to consider whether such models could serve as cognitive models of the word-by-word difficulty of processing a sentence in humans \citep{hagoort_neural_2019, linzen_syntactic_2021, engelmann_processing_2009, arehalli_neural_2024, schrimpf_neural_2021, caucheteux_brains_2022, goldstein_shared_2022}. 

Evaluating this possibility requires a linking hypothesis between some measurement derived from the language model (LM) on the one hand and a behavioral measure of word-by-word processing difficulty in humans on the other hand. One such linking hypothesis is provided by surprisal theory \citep{hale_probabilistic_2001, levy_expectation-based_2008}, which argues that the difficulty of processing each word in a sentence--measured, for example, as the time a reader spends fixating on the word--is proportional to the \textit{surprisal} of that word in context (i.e. its negative log probability). The strong version of this hypothesis holds that all sources of processing difficulty---temporary ambiguity, syntactic complexity, word frequency---can be reduced to word surprisal. This hypothesis can be fruitfully connected to LMs: If we can identify an LM whose surprisal estimates can consistently explain reading difficulty, this would suggest both that surprisal theory provides a good characterization of sentence comprehension at Marr's computational level \citep{marr_vision_1982}, and beyond this, that the representations and constraints underlying the LM's predictions may parallel those underlying the same process in humans \citep[e.g.][]{hale_finding_2018, ryu_memory_2025}.

\begin{figure*}[tb]
\includegraphics[width=\columnwidth]{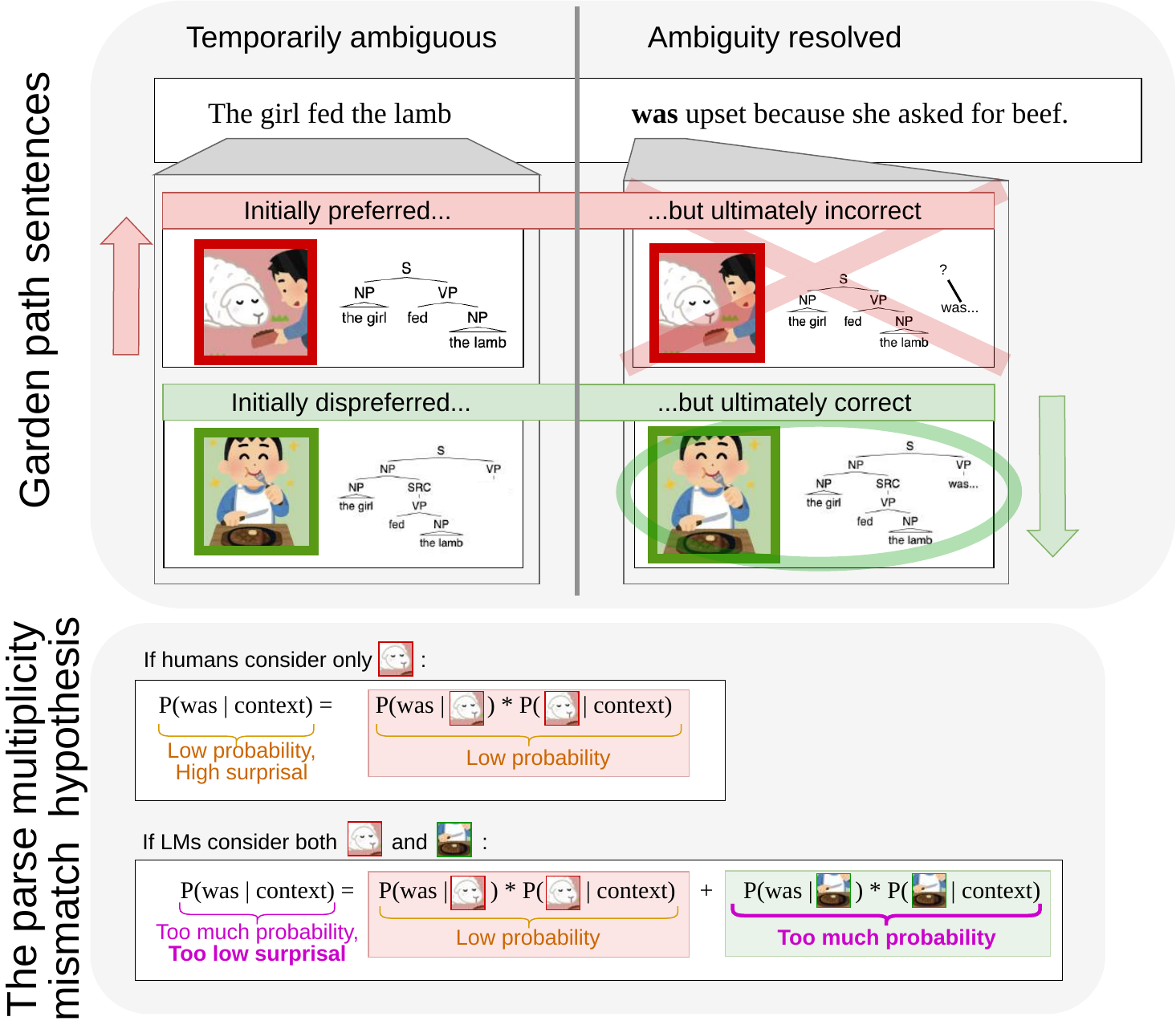}
\caption{(Top) An example of a garden path sentence with its initially preferred and globally correct interpretations. Humans experience processing difficulty when the ambiguity is resolved. (Bottom) The hypothesized relationship between parse multiplicity and the surprisal of the disambiguating word. (Illustrations courtesy of  \href{https://www.irasutoya.com/}{Irasutoya})}
\label{fig:fig1}
\end{figure*}

Empirically, LM surprisal has had mixed success predicting human reading behavior. There is a robust correlation between LM surprisals and human reading times on naturalistic texts \citep{smith_effect_2013,wilcox_predictive_2020, shain_large-scale_2024}, and across languages \citep{wilcox_testing_2023}. But correlational studies alone are insufficient to establish the validity of surprisal theory as a general theory of processing difficulty, and LM surprisal has fared less well at predicting human reading behavior in controlled experiments designed to adjudicate between sentence processing theories \citep{slaats2026more}. Misalignment between LMs and humans is particularly drastic in \textit{garden path sentences}--temporarily ambiguous sentences that initially favor one interpretation, but are ultimately disambiguated in favor of a dispreferred interpretation (Figure \ref{fig:fig1}, top) \citep{van_schijndel_single-stage_2021, wilcox_targeted_2021, arehalli_syntactic_2022, kobzeva2024grammar, huang_large-scale_2024}. Garden path sentences have long served as a crucial testbed for understanding how comprehenders deal with the pervasive problem of linguistic ambiguity during online processing. For example, consider an utterance beginning with ``The girl fed the lamb...''. This utterance is consistent with at least two interpretations: A preferred interpretation, in which a girl is feeding the lamb; and a dispreferred interpretation, in which the girl is being fed the lamb. Readers experience significant processing difficulty in cases when the sentence is disambiguated in favor of this dispreferred interpretation (e.g. if it is continued with ``...was upset because she asked for beef.'', which is an ungrammatical continuation under the initially preferred reading, but perfectly grammatical under the initially dispreferred reading). The disambiguation cost can be isolated by measuring reading times starting at the disambiguating word ``was'' in the temporarily ambiguous sentence \ref{ex:ambig_full}, and comparing this to reading times on the same word in an unambiguous control sentence \ref{ex:unambig_full}. The difference in reading times between the two conditions is referred to as a \textit{garden path effect}:

\begin{enumerate}[resume, noitemsep]%
    \item{The girl \phantom{\textbf{who was}} fed the lamb \phantom{.........} \textit{was} upset because she asked for beef. (temporarily ambiguous) \label{ex:ambig_full}}
    \item{The girl \textbf{who was} fed the lamb \phantom{.........} \textit{was} upset because she asked for beef. (unambiguous) \label{ex:unambig_full}}
\end{enumerate}

In a self-paced reading experiment, \cite{huang_large-scale_2024} shows that the entire disambiguating region in this construction (the disambiguating word and two following words---the difficulty incurred by a given word often manifests at the following words as well) takes more than 300ms longer to read in the ambiguous condition compared to the unambiguous condition. LMs have been shown to predict the direction of these garden path effects: For example, the GPT-2 language model \citep{radford2019language} estimates surprisal of the disambiguating region to be about 4 bits higher on average in the ambiguous condition than the unambiguous condition for this construction \citep{huang_large-scale_2024}. However, this surprisal difference fails to explain the full \textit{magnitude} of the effect. Simplifying the modeling details, \cite{huang_large-scale_2024} estimate that reading times increase by roughly two milliseconds for every one bit increase in GPT-2 surprisal. If this relationship is assumed to also hold in garden path sentences, as argued by surprisal theory, then GPT-2 surprisal only predicts a garden path effect of approximately 8ms (4 bits * 2ms/bit), nearly 40 times smaller than the empirically observed effect \citep{huang_large-scale_2024}. This means garden path effects cannot be reduced to a difference in surprisal from any the language models tested. In a separate study, \citet{timkey_eye_2025} show this pattern holds across a diverse range LM architectures, training objectives and training corpora.

There are at least two logically possible reasons for these misalignments between LMs and humans. It may be the case that surprisal theory is wrong, and the processing difficulty of a word cannot be fully reduced to its predictability. Another possibility---the focus of the present work---is that surprisal theory is generally correct, but LMs and humans systematically diverge in their expectations about upcoming words because humans are more computationally constrained than LMs. If this is the case, the alignment between human behavior and LM next-word predictions could be improved by imposing more humanlike resource constraints on LMs \citep{timkey_language_2023,clark_linear_2025,oh_model_2025}. 

In this work, we investigate whether the gap between humans and LM predictions in garden path sentences could be explained by differences in their \textit{parse multiplicity}: The number of distinct \textit{interpretations} (for our purposes, distinct syntactic parses) of an utterance that can be maintained simultaneously in memory.\footnote{We use the term ``multiplicity'' over another commonly used term ``parallelism'', to avoid confusion with another unrelated sense of the latter, which referring to the number of parallel syntactic computations a parser can execute, rather than the number of pareses that can be simultaneously maintained \cite{lewis_falsifying_2000}.} We refer to this hypothesis as the \emph{parse multiplicity mismatch hypothesis}. Concretely, we expect the magnitude of garden path effects to increase as fewer interpretations can be considered. For example, if a comprehender must commit to only a single interpretation of the utterance ``the girl fed the lamb'', they will likely commit to the more frequent interpretation of the girl giving food to the lamb. Under this interpretation, the disambiguating continuation of ``was'' is not only unexpected, but potentially ungrammatical (cf. ``The girl gave the lamb was upset...''). By contrast, when more interpretations are considered simultaneously when generating expectations about the next word, the chances increase that at least one of these interpretations is the globally correct one. Under the globally correct interpretation, in which the girl is being fed the lamb, ``was'' is a grammatical, predictable continuation (cf. ``The girl given the lamb was upset...''). This increases the overall predictability of the word ``hates'', decreasing the magnitude of the garden path effect (Figure~\ref{fig:fig1}, bottom). 

The degree of parse multiplicity of both humans and LMs remains poorly understood. In humans, this has been a major topic of contention for decades \citep{boland1996interaction, gibson_distinguishing_2000, lewis_falsifying_2000, clifton2008parallelism}. Some incremental parsing models assume a fully serial architecture, in which only a single parse is maintained at once \citep[e.g.][]{frazier_sausage_1978, marcus1978theory, frazier_sentence_1987, gorrell1995syntax, pritchett1988garden, lewis_reanalysis_1998, clifton_specifying_1999, van2005evidence, lewis_activation-based_2005}, while others have argued that multiple distinct syntactic hypotheses can be maintained in parallel \citep[e.g.][]{gibson_computational_1991, hale_probabilistic_2001, levy_expectation-based_2008, trueswell_toward_1994}. \cite{jurafsky_probabilistic_1996} explicitly argues that garden path effects arise from a fixed limit on parse multiplicity in humans, and treats this limit as a free parameter of the parsing model to be estimated from reading data \citep[see also][]{boston_parallel_2011}. By contrast, evidence suggests that LMs, which are typically not designed with explicit syntactic representations, can nevertheless implicitly represent and maintain multiple syntactic parses of a sentence in parallel \citep{aina_language_2021,eisape_probing_2022,hanna_incremental_2024}. Mathematically, an LM's ability to accurately approximate the probability distribution over next word of a sentence should strictly improve with greater parse multiplicity: The most accurate estimate of next-word probability results from marginalizing, implicitly or explicitly, over all possible parses of the sentence---that is, taking a weighted average of the next word's probability under each possible parse, where the weights are determined by the probability of the parse. \textbf{If LMs do in fact have greater parse multiplicity than humans, then reducing their parse multiplicity should lead to more human-like predictions in garden path sentences.}

\begin{figure*}[tb]
\includegraphics[width=\columnwidth]{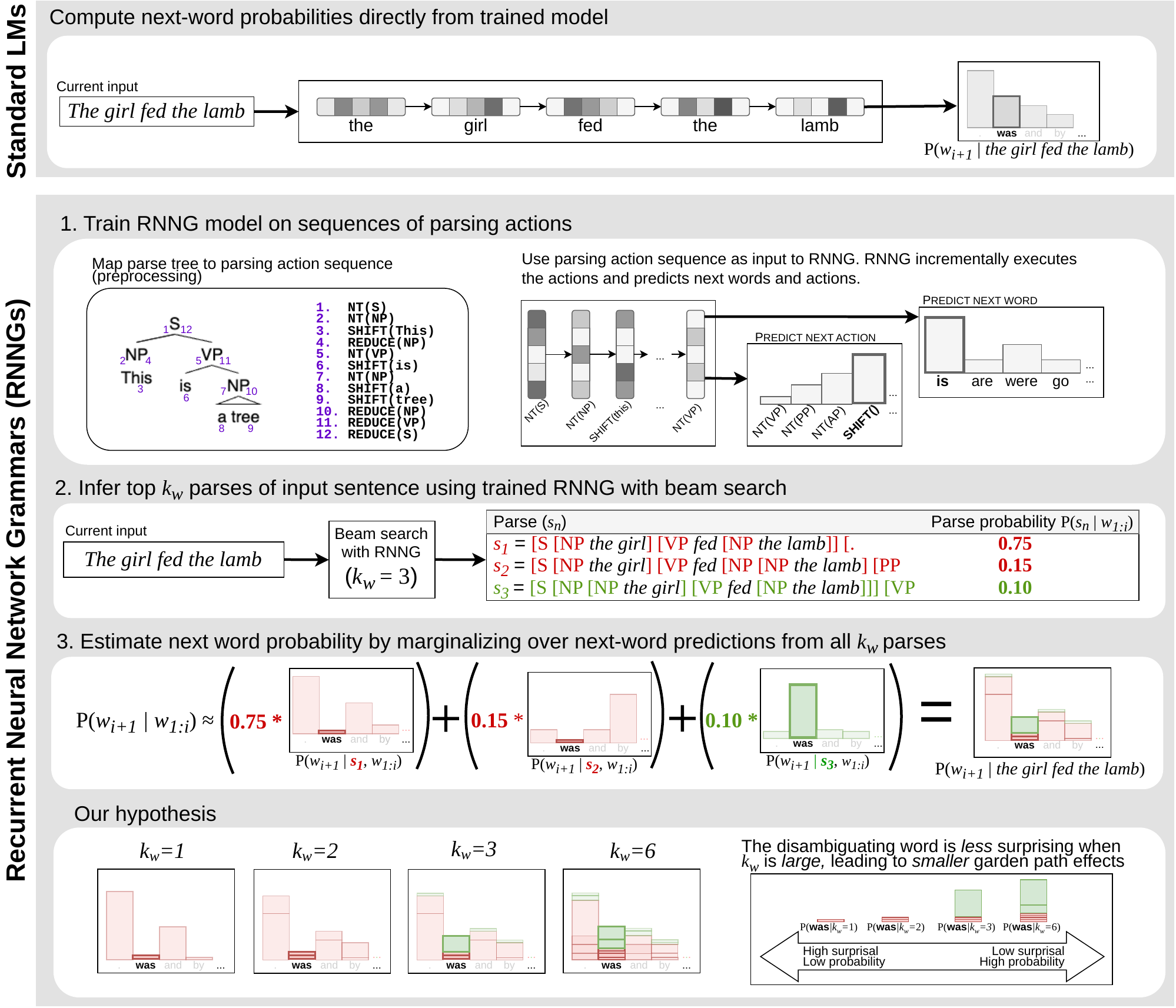} 
\caption{Comparing the computation of next word probabilities in standard LMs, and in RNNGs. This example demonstrates the top-down parsing strategy.}
\label{fig:rnng_vs_lm}
\end{figure*}

One challenge in evaluating the parse multiplicity mismatch hypothesis is that parse multiplicity cannot be directly manipulated in the most commonly used LMs such as GPT-2, as they are trained to predict the next word conditional only on the previous words. Any syntactic representations they have are emergent and implicit, making it difficult both to understand which interpretations an LM is considering, and to intervene on these implicit representations \citep[though see][]{hanna_incremental_2024}. Here, we address this issue by using Recurrent Neural Network Grammars (RNNGs; \citealt{dyer_recurrent_2016}), models trained to predict both the words and structure of a sentence. These models condition their next word predictions on an explicit syntactic parse of the context. From these models, we can infer the $k$ most likely parses of the previous words using a variant of the beam search algorithm (to be described below), and then approximate next-word surprisals by marginalizing over these $k$ parses (Figure \ref{fig:rnng_vs_lm}). In this setup, we can systematically vary the parse multiplicity parameter $k$ (commonly referred as the beam width), which determines the number of distinct parses maintained in parallel. A fully serial parser corresponds to $k=1$, and a fully parallel parser is approximated as $k$ approaches infinity. 

We test whether reducing parse multiplicity in RNNG-based LMs leads to larger predicted garden path effects, and whether such increases are sufficient to explain the full magnitude of garden path effects in humans. This hypothesis is only partially supported: We find that models with less parallelism do predict larger garden path effects, but even the largest predicted effects are still orders of magnitude smaller than those observed in humans. This pattern of results holds even when any parses consistent with the globally correct interpretation are manually excluded from the parser's set of hypotheses (in other words, when the models are forced ``down the garden path''). Our results suggest that human syntactic disambiguation difficulty cannot be reduced to word surprisal from an RNNG-based LM, regardless of the number of interpretations that can be entertained at once. The finding that surprisals derived from a serial parser forced down the garden path cannot capture the magnitude of syntactic diambiguation in reading suggests that explaining the full magnitude of garden path effects will require positing additional mechanisms, such as syntactic reanalysis.\footnote{Data and analysis code are provided at \href{https://osf.io/7mprn/}{https://osf.io/7mprn/}}

\section{Materials}
\label{subsec:materials}
Materials for all experiments come from the garden path subset of the SAP Benchmark \citep{huang_large-scale_2024}. This subset of the dataset contains self-paced reading data from 2000 participants on three different types of garden path constructions (in the self-paced reading paradigm, participants are presented a sentence one word at a time, pressing the space bar to reveal each new word, after which the previous words are replaced with dashes; The time between keypresses is recorded as the word's reading time). There are two versions of each sentence, one with a temporary ambiguity, and one where extra material (here indicated in parentheses) has been added to eliminate this temporary ambiguity.\\

\begin{minipage}{\linewidth}
\noindent \textbf{MV/RR}: The girl (who was) fed the lamb \textit{remained} calm despite having asked for beef. \\
\noindent \textbf{NP/S}: The girl found (that) the lamb \textit{remained} calm despite the absence of its mother.\\
\noindent \textbf{NP/Z}: When the girl attacked(,) the lamb \textit{remained} calm despite the sudden assault.
\end{minipage}\\

In the MV/RR construction, the verb `fed' is initially ambiguous between being the main verb (MV; initially preferred), or an embedded verb in a reduced relative clause modifying `the girl' (RR; globally correct). The actual main verb, `remained', disambiguates the sentence toward the RR interpretation. When the words `who was' are added, `fed' is unambiguously part of a relative clause. 

The NP/S ambiguity is caused by the fact that the verb `found' can either take a noun phrase (NP; initially preferred), or a sentential (S; globally correct) complement. The relativized verb `remained' disambiguates the sentence in favor of the sentential complement interpretation. With the addition of the complementizer `that', the following noun phrase, `the lamb', is unambiguously the subject of a sentential complement, rather than the direct object of `found'.

In the NP/Z ambiguity, the verb `attacked' can optionally take a noun phrase as complement (NP; initially preferred), or no complement (Z; globally correct). The relativized verb `remained' disambiguates the sentence in favor of the zero complement (Z) interpretation. When a comma is added after `attacked', this signifies the end of the clause, meaning the verb unambiguously takes zero complement, and that `the lamb' is part of a new clause.

\cite{huang_large-scale_2024} compare reading times between the two versions of each sentence in the disambiguating word, ``remained'', and the two words that immediately follow, ``calm'' and ``despite'', which are referred to as the first spillover and second spillover words respectively. Here, we focus on the first spillover word, which displays the largest empirical effects \cite{huang_large-scale_2024}. Results from the other two words, which did not differ qualitatively from the results at the first spillover word, are reported in Appendix~\ref{sec:bllip_results}.

\section{Methods}
\subsection{Recurrent Neural Network Grammars}
\label{sec:rnngintro}
Standard language models are trained to predict the next word in a sequence given the previous words as input, $P(w_{i+1} | w_{1:i})$. By contrast, RNNGs jointly predict the next word and next parsing step given a particular syntactic parse of the previous words. The probability of a parse is estimated through the sequence of actions a transition-based constituency parser would take to produce the parse. Simplifying somewhat, these actions can be either structural, where words are composed into phrases (i.e. \texttt{NT(X)}, which predicts a nonterminal, and \texttt{REDUCE(X)}); or lexical, where the next word is predicted and then focus shifts to that word (the \texttt{SHIFT} action). Given the sequence of all $i$ previous words, $w_{1:i}$, and all $j$ previous parsing actions $s_{1:j}$ that form the partial derivation of the parse tree up until just before $w_{i+1}$, the RNNG jointly predicts the next parsing action $P(s_{j+1}| s_{1:j},w_{1:i})$, and the next word, $P(w_{i+1}|s_{1:j} w_{1:i})$. The next word distribution is only defined when the next action is \texttt{SHIFT}, as words can only be added to the parse in tandem with a \texttt{SHIFT} action. 

\subsection{Varying parse multiplicity in RNNGs using word-synchronous beam search}
The ultimate quantity of interest for the present study is a word's surprisal given all previous words, $P(w_{i+1} | w_{1:i})$. Unlike standard language models, which estimate this quantity directly, RNNGs condition next word predictions on a parse of the previous words, $P(w_{i+1} | s, w_{1:i})$. To produce exact next-word probabilities we would need to marginalize over the set of all possible parses of the previous words, $S$---that is, average the parse-conditional next-word distributions, weighted by each parse's probability:
\begin{equation}
  P(w_{i+1} | w_{1:i}) = \sum_{s \in S}^{} P(w_{i+1} | s,w_{1:i})P(s| w_{1:i})
\end{equation}

\noindent The marginalization required to compute $P(w_{i+1} | w_{1:i})$ from an RNNG cannot be performed exactly in practice, as $S$ quickly becomes far too large: Even short sentences in simple grammars are generally compatible with many different partial parses. But this quantity can be well approximated by first using an inference algorithm to search for the top $k$ most likely parses of $w_{1:i}$, and then marginalizing the next word distribution over only those $k$ parses (Figure \ref{fig:rnng_vs_lm}). In this work, we perform inference on our RNNG models using a beam search algorithm \citep{lowerre_harpy_1976}, specifically the \citet{noji_effective_2021} implementation of \textit{word-synchronous beam search} \citep{stern_effective_2017}. At a high level, this procedure incrementally constructs a set of candidate parses for the input by iteratively attempting sequences of parsing actions, and retaining only a specified number of the most probable parses at each word. The key parameter governing this process is the \textit{word beam width}, $k_w$, which specifies how many distinct parses are retained as each word is observed. The term ``beam width'' comes from the metaphor of searching in the dark with a flashlight: The width of the light beam represents the small set of possibilities that are visible at a given time. With a large beam width, the model can maintain many competing interpretations of the input, more closely approximating full marginalization over possible parses. With a small beam width, the model's next word predictions can only reflect a small number of analyses. In the present work, we compute surprisals using $k_w = \{1,2,3,4,5,10,25,50,100,250,500,1000\}$. 

Word-synchronous beam search also includes a second hyperparameter, the \textit{action beam width} $k_a$ (typically denoted simply as $k$), which we fix to a large value of $k_a$ = 1000 for all $k_w$. We discuss the motivations for this decision in Appendix \ref{sec:action_beam}.

\subsection{Upper and lower bounds through counterfactual beam manipulations} The parse multiplicity mismatch hypothesis---in the terms of the previous section, the hypothesis that narrower beams will lead to larger and therefore more humanlike garden path effects---relies on two premises. The first premise is that by decreasing the beam size, we increase the chances that parses consistent with the globally correct interpretation are pruned from the beam. The second is that the surprisal of the disambiguating word under the initial interpretation, which remains on the beam, is at least large enough to capture the full magnitude of the human garden path effect magnitude; the logic here is that if the globally correct interpretation, which is consistent with the disambiguating word, were still on the beam, the total surprisal of the word (marginalized over the two interpretations) would, if anything, be even lower, which in turn might account for LMs' inability to model the magnitude of garden path disambiguation difficulty. This entails that the disambiguating word's surprisal conditioned on the initial, incorrect interpretation alone serves as an upper bound on the predicted garden path effect size that we can expect to derive from RNNGs surprisal. 

This upper bound is important for evaluating the parse multiplicity hypothesies, but note that it does not necessarily correspond to the disambiguating word's surprisal when $k_w = 1$. Multiple parses of the ambiguous region are consistent with the initially preferred interpretation. For example, the incremental parses $s_1$ and $s_2$ in Figure \ref{fig:rnng_vs_lm} differ only in whether they posit prepositional modifier following ``the lamb'', but they do not differ with respect to the ambiguity of interest; both treat ``fed'' as a main verb (MV) instead of a reduced relative (RR) verb as in $s_3$. Thus, if we want to calculate the surprisal of the disambiguating word given a particular interpretation, we must marginalize over all parses that are consistent with that interpretation. To this end, in addition to surprisals from the various fixed $k_w$ values, we also compute these upper-bound surprisals in what we refer to as the \textsc{Forced Garden Path} condition: We select the largest beam width $k_w = 1000$, manually sort the parses into bins according to the ambiguity of interest (i.e. whether ``fed'' is a main verb or part of a reduced relative clause), and then compute surprisal by marginalizing over only the parses in the bin that correspond to the main verb interpretation, instead of over all 1000 parses. This is done only for the ambiguous sentences; for the unambiguous versions of the sentences, we simply use the unmodified surprisal estimates at $k_w = 1000$.

At the other extreme, it is still possible---and in fact quite common---that even at $k_w = 1000$ all parses are only consistent with the initial interpretation, and no parses are consistent with the globally correct one. To address this issue, we introduce a lower-bound ``fully-parallel'' condition designed to more explicitly approximate a fully parallel parser. This condition guarantees that at least one parse consistent with the globally correct interpretation is always on the beam before and after disambiguation. Our method for manually constructing the beam and computing surprisals in this condition is described in Appendix~\ref{sec:full_parallel_cond}.

\subsection{Top-down and Left-corner parsing strategies}
\label{sec:parsing_strats}
We evaluate RNNGs trained with two different parsing strategies, top-down and left-corner parsing. Since the parse multiplicity mismatch hypothesis does not favor one of these strategies over the other, we experiment with both.

Parsing strategies differ from one another in how eager they are to make predictions about a sentence's structure. Bottom-up parsing is minimally predictive: It waits until every word of a constituent is observed before positing the constituent in the parse (we do not experiment with this strategy). Top-down parsing is maximally predictive: It eagerly posits every node dominating the next word in the tree as soon as the current word is observed. The left-corner strategy falls in the middle of these two strategies. A left-corner parser waits until a word is observed, then eagerly predicts only its parent and sister node. For example, given the noun phrase (NP) ``The girl'', a bottom up strategy predicts the NP node after observing ``girl'', a left corner strategy does so after observing ``the'' and a top-down strategy does so before observing ``the''.

Left-corner parsers have been argued to be more psychologically plausible than top-down and bottom-up parsers \citep{resnik_left-corner_1992, nelson2017neurophysiological}. Left-corner parsers are also more memory efficient than top-down parsers, requiring a much smaller beam size to achieve similar performance to top-down parsers \citep{kuncoro_lstms_2018}. We conduct our experiments using both top-down and left-corner RNNG models, but we do not intend to adjudicate between the two in the present work. We do not test models with the minimally predictive bottom-up strategy as their incremental parses do not distinguish between the two interpretations prior to disambiguation in both the MV/RR and NP/S constructions.

\subsection{Model training} We use the \citet{noji_effective_2021} RNNG implementation in pytorch. RNNGs need to be trained on syntactic trees; this implementation preprocesses each tree into a sequence of left-corner or top-down parsing actions. These trees can come from a hand-annotated treebank, such as the Penn Treebank \citep{marcus_building_1993}, or a machine-parsed corpus. We opt for latter to avoid sparsity issues caused by the limited size of hand-annotated treebanks. We use two corpora for model training: A 42M word subset of the BLLIP corpus, commonly used to train RNNG models in prior works \citep{hu_systematic_2020} and the 100M word BabyLM corpus \citep{warstadt_findings_2023}, a corpus of child directed speech, television subtitles, stories, and Wikipedia articles, designed to approximate the quality and quantity of linguistic input a child receives before adulthood. Both corpora were parsed using the Berkeley parser \citep{kitaev_constituency_2018}. For all four combinations of datasets and parsing strategies, we train five random seeds of the RNNG models, giving us 20 models in total. 

To ensure that our trained models provide high quality estimates of parse probabilities when combined with beam search, we evaluated their parsing performance on the WSJ subset of the Penn Treebank \citep{marcus_building_1993}. Their performance was surprisingly competitive with state-of-the-art constituency parsers (see Appendix~\ref{sec:parsing_performance}). In what follows, we focus on models trained on the BabyLM corpus, as these models consistently attained better parsing performance on the Penn Treebank than the models trained on the BLLIP corpus, but results were remarkably similar across both training datasets (results from the BLLIP models are summarized in Appendix~\ref{sec:bllip_results}).

\subsection{Predicting reading times from surprisal}
\label{sec:surp_to_rt}
To predict reading times from RNNG surprisals, we follow the methodology established by \cite{van_schijndel_single-stage_2021}, and used in a number of studies since then \citep[e.g.][]{wilcox_targeted_2021, arehalli_syntactic_2022,huang_large-scale_2024,kobzeva2024grammar}. In the \cite{huang_large-scale_2024} study, the garden path sentences read by participants  were interleaved with naturalistic filler sentences. We use reading data collected on these filler sentences to estimate the relationship between word surprisal and reading times, as follows. We fit linear mixed-effect regression models to predict reading times from model surprisal estimates and several nuisance variables (word length, frequency, position) on the filler sentences. Then, the fitted model is used to predict reading times in the garden path stimuli. When converting surprisals to reading times in the \textsc{Forced Garden Path} and \textsc{Full-Parallel} conditions, we use filler models fit to the $k_w = 1000$ surprisal estimates. Estimating the surprisal-to-reading time relationship using only filler sentences tests a key prediction of surprisal theory: If surprisal can fully explain processing difficulty in complex and simple sentences alike, then the relationship between surprisal and processing difficulty in the filler sentences should generalize to all sentences, including garden paths \citep{van_schijndel_single-stage_2021}. We then estimate the difference in predicted reading times between the ambiguous and unambiguous conditions at the disambiguating word and the two following words (to account for spillover effects) using Bayesian linear mixed effects regression models. This process is repeated for all 12 values of $k_w$ for each of the 20 RNNG models. Our statistical methods are described in more detail in Appendix~\ref{sec:regression_info}.

\begin{figure*}[tb]
\includegraphics[width=\columnwidth]{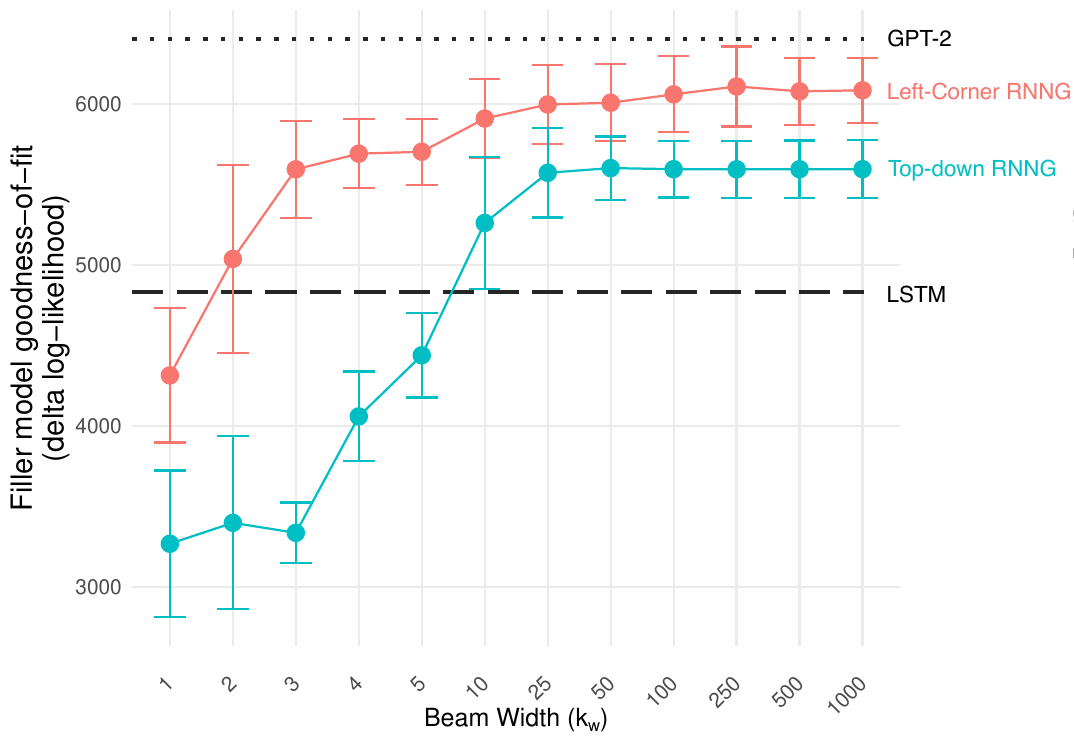} 
\caption{Goodness of fit (increase in log-likelihood compared to baseline model; higher is better) to filler sentences improves when surprisal is estimated using larger beam widths. Results from the language models investigated in \cite{huang_large-scale_2024}---GPT-2 small \citep{radford2019language} and an LSTM language model \citep{gulordava_colorless_2018}---are provided for reference. Error bars are standard errors over each condition's 5 model seeds.}
\label{fig:filler_dll}
\end{figure*}

\begin{figure*}[tb]
\includegraphics[width=\columnwidth]{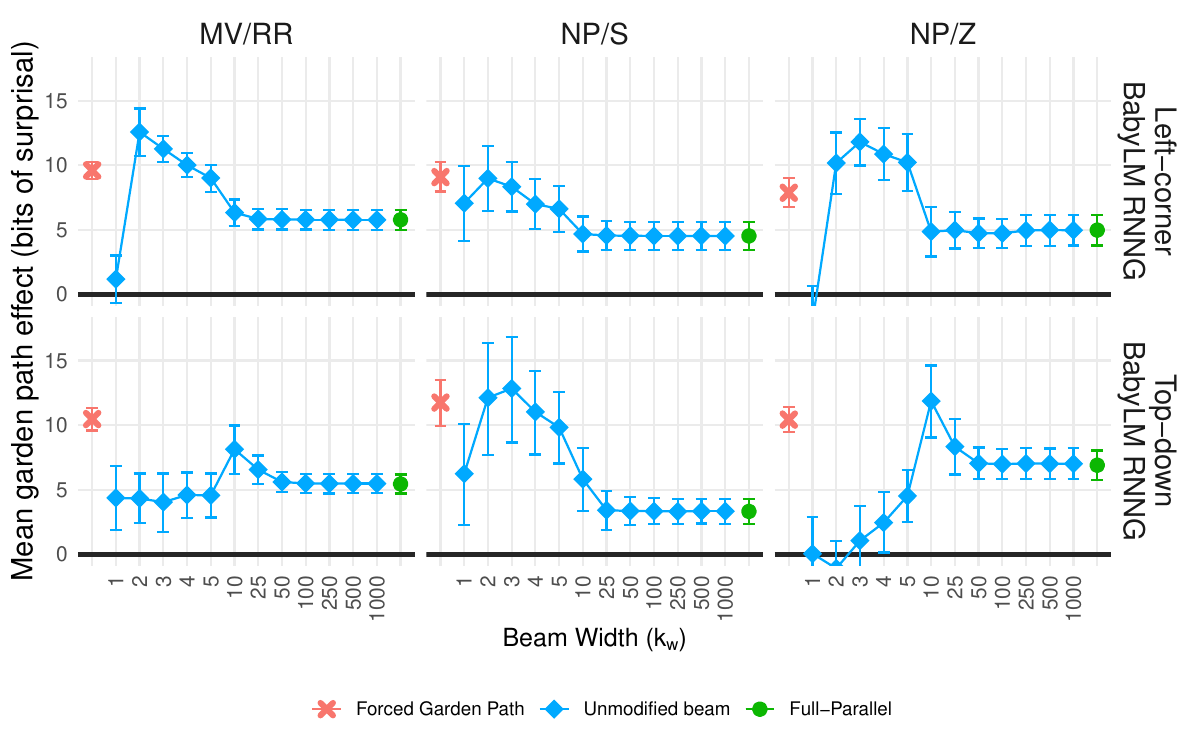} 
\caption{Predicted garden path effects summed across the disambiguating word and two spillover words, measured in surprisal, by construction type, parsing strategy and beam width, including the two counterfactual beam conditions---the \textsc{Forced Garden Path} condition, which serves as an upper bound on the effect, and the \textsc{Full-Parallel} lower bound---and two parsing strategies. Error bars represent 95\% credible intervals, estimated from Bayesian regression models, described in Appendix~\ref{sec:regression_info}.}
\label{fig:gpes_by_beamsize_surp}
\end{figure*}

\begin{figure*}[tb]
\includegraphics[width=\columnwidth]{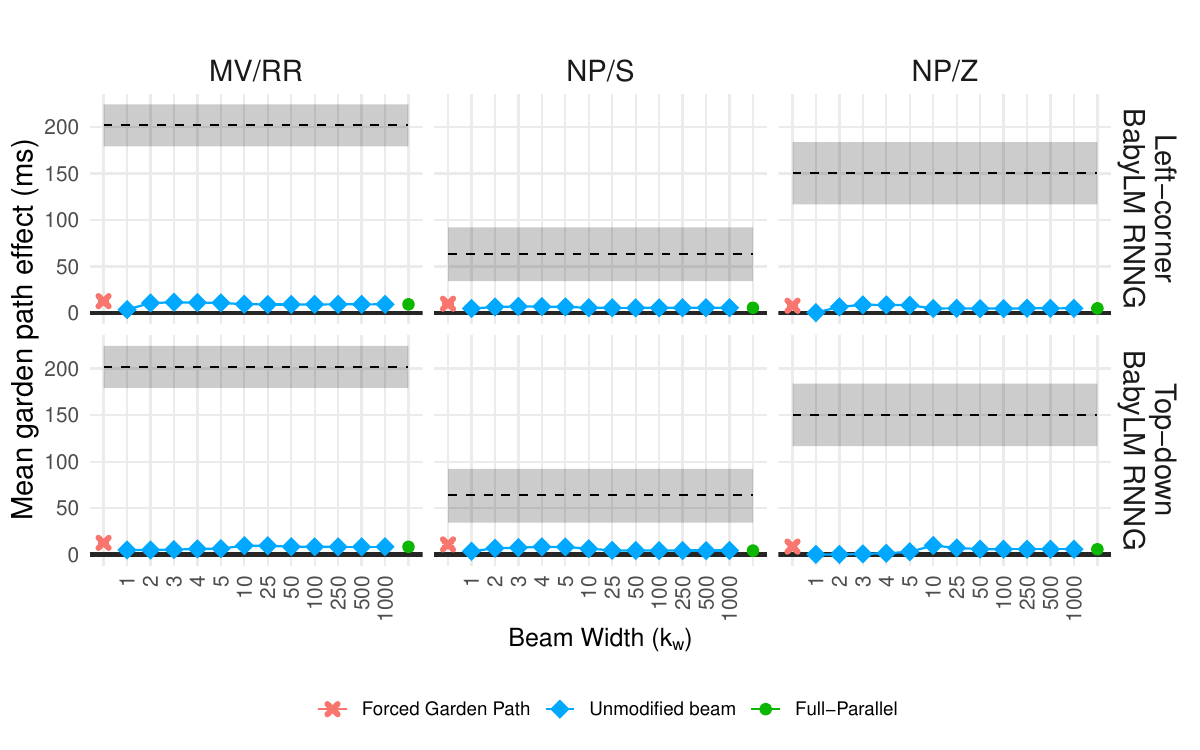} 
\caption{Predicted garden path effects at the first spillover region, measured in milliseconds of reading time, across the three construction types, 12 beam sizes (1-1000) and the two counterfactual beam conditions \textsc{Forced Garden Path} and \textsc{Full-Parallel}, and two parsing strategies. Empirical effect sizes are represented as horizontal bands. Error bars represent 95\% credible intervals, estimated from Bayesian regression models, described in Appendix~\ref{sec:regression_info}.}
\label{fig:gpes_by_beamsize_rts}
\end{figure*}

\section{Surprisal better predicts reading times in naturalistic sentences as syntactic parallelism increases}
Before turning to the garden path sentences, we first examine the general relationship between parse multiplicity and the ability of surprisal to predict reading times in the naturalistic filler sentences. We quantify gooddness of fit as the increase in log-likelihood on the filler data of a regression model including surprisal predictors over a model with only the baseline predictors of word length, frequency and position (see Appendix~\ref{sec:regression_info} for details). Each regression model contains surprisal estimates at only a single $k_w$.

We find that surprisal provides a better fit to filler sentence reading times at larger beam widths, as shown in Figure \ref{fig:filler_dll}. RNNGs with a left-corner strategy show a consistently better fit than the top-down models across all beam widths, but especially at smaller beam sizes, consistent with prior results indicating that left-corner parsers require smaller beam sizes to reach comparable parsing performance to top-down parsers \citep{noji_effective_2021}. Improvements in goodness-of-fit quickly diminish in both strategies around $k_w = 25$. These results are consistent with a range of past work predicting reading times in naturalistic sentences from surprisal across various beam widths, which have shown that larger beam sizes result in consistently better fits to human reading times \citep{boston_parallel_2011, hale_finding_2018, noji_effective_2021}.

\section{Constraining parallelism leads to larger (but not large enough) garden path effects}
\subsection{Setup}
In this section, we assess the relationship between parse multiplicity and the size of predicted garden path effects. First, we report the magnitude of garden path effects in bits of surprisal across various beam sizes. Then, we use the methods outlined in Section~\ref{sec:surp_to_rt} to predict garden path effect sizes in reading times using the surprisal-to-reading time conversion factors estimated from the filler sentences.

\subsection{Limited parse multiplicity increases surprisal}
Across all constructions and parsing strategies, predicted garden path effects were largest at smaller-to-intermediate beam widths, consistent with the parse multiplicity mismatch hypothesis (Figure~\ref{fig:gpes_by_beamsize_surp}). The effect always decreased in magnitude before reaching an asymptote at the largest beam widths. Interestingly, garden path effects were consistently smallest at $k_w=1$, which seems to contradict our hypothesis that garden path effect magnitudes will increases as $k_w$ decreases. A manual inspection of the beam revealed that parses were generally of very poor quality at $k_w = 1$ in both the ambiguous and unambiguous conditions, making the disambiguating word an equally poor continuation with similarly high surprisals in both conditions. There appears to be a ``sweet spot'' of $2 \leq k_w \leq 10$ where the beam width is large enough to produce parses of reasonable quality in both conditions, but small enough that the globally correct parse is pruned from the beam in the ambiguous condition before the disambiguating word is observed. The results from the \textsc{Forced Garden Path} and \textsc{Full-Parallel} conditions are also consistent with our hypothesis: Garden path effects were consistently larger when models were explicitly forced to consider only parses consistent with the initial interpretation than when they are forced to always consider both interpretations. 

This result appears to be inconsistent with our observation that larger beam widths provided better fits to human reading times in the filler sentences, if the parse multiplicity mismatch hypothesis is expected to explain the gap between reading times predicted from LM surprisals and empirical reading times. If the number of parses one can maintain in parallel is a fixed cognitive constraint that does not depend on the particular properties of the sentence---as assumed in prior models such as \citet{jurafsky_probabilistic_1996}---then the beam width that provides the best fit to reading times in filler sentences should also predict the largest (i.e. most humanlike) garden path effects. 

To explicitly evaluate whether decreased beam size is associated with an increased probability of the globally correct interpretation being pruned from the beam, we manually inspected the relative probability of the initially preferred and globally correct interpretations at the three words preceding and following disambiguation. Detailed results can be found in Appendix \ref{sec:interp_probs}. We indeed find that the relative probability of the globally correct interpretation is lower at smaller beam widths than larger beam widths prior to disambiguation.

\subsection{Limited parse multiplicity does not close the gap with reading times}
As shown in Figure \ref{fig:gpes_by_beamsize_rts}, surprisals at all beam widths drastically underestimate the magnitude of the effects observed in humans in both parsing strategies and all three constructions. This suggests that the misalignment of language models and human behavior in garden path sentences cannot be alleviated through limits on parse multiplicity alone. The qualitative similarity of results in the two extreme conditions \textsc{Forced Garden Path} and \textsc{Full-Parallel}, shows that the underprediction of garden path effects cannot be attributed to simple differences between models and humans in the parses under consideration prior to disambiguation: Effects are underpredicted by orders of magnitude both when no globally correct parse is under consideration, and when globally correct parses are guaranteed to be under consideration.

\section{Discussion}
The present work evaluates the parse multiplicity mismatch hypothesis, according to which the misalignment between language model predictions and human reading behavior in garden path sentences is caused by a misalignment in the computational constraints of neural language models and humans, and in particular, a misalignment in the number of simultaneous syntactic analyses of an utterance that can be maintained in memory. This hypothesis would be a way to reconcile the strong version of the surprisal hypothesis with the large empirical effects observed in human studies: If word surprisal estimates from a memory-limited language model can predict the full magnitude of garden path effects, then this would suggest both that sentence processing difficulty can indeed be reduced to word predictability, and that the mechanisms underlying this process in humans are constrained in a similar way \citep{oh_model_2025}. 

We have operationalized this memory limitation using Recurrent Neural Network Grammars combined with the word-synchronous beam search inference algorithm, which together can provide word predictability estimates given any specified number $k_w$ of parallel candidate syntactic parses. While we found robust evidence that RNNGs predict larger garden path effects when we reduce their syntactic parallelism (lower $k_w$), the predicted effects were dwarfed by the empirical ones for all values of $k_w$, suggesting that this hypothesized memory constraint cannot close the gap between LM surprisals and human reading behavior.

Our results are unlikely to depend on our choice to use the word synchronous beam search inference algorithm over alternatives such as particle filters \citep{levy_modeling_2008, maina-kilaas_algorithmic_2026}, or beam search with different pruning criteria, such as the relative-width beam criterion proposed by \cite{jurafsky_probabilistic_1996}. Even in the \textsc{Forced Garden Path} condition, where surprisals were computed using only parses consistent with the initially preferred interpretation, predicted effects remained far too small. This suggests that the misalignment is not driven by the subset of parses considered by the model, but by the too-low surprisal assigned to the disambiguating word \textit{given the initially preferred interpretation} (an interpretation that renders the word ungrammatical). This problem would persist if we used another algorithm to perform inference over parses, as this choice can only influence which parses are under consideration, not the surprisal of the next word given a particular parse.

\subsection{Reanalysis}
One family of parsing models assumes that parsing proceeds in two stages \citep{frazier_sausage_1978, lewis_reanalysis_1998}. In the first stage, readers incrementally construct a limited set of possible candidate parses for the utterance. This is followed by a second stage, an error-driven repair process occurring when the next word of a sentence cannot be integrated with any of the candidate parses. In the present work, we implicitly assumed a single-stage account of syntactic disambiguation difficulty, in which there are no separate reanalysis mechanisms, and all of the costs associated with processing any word---including the disambiguating word of a garden path sentence---should be reflected in the word's surprisal. 

Our results suggest that a single-stage account based on language model surprisal is not sufficient, regardless of memory constraints, but it may still be possible for limited parallel language models to provide a partial account of disambiguation difficulty in a two-stage architecture. A two-stage account of processing has at least three requirements: A means of computing ``stage one'' parsing costs, a means of detecting parsing failures in order to activate the second stage, and a means of computing ``stage two'' reanalysis costs. Predictability estimates from RNNGs with beam search may be able to fulfill the first two of these three requirements. For example, surprisal from a limited parallel model might provide a generally sufficient characterization of the costs of processing a word when there is no integration failure, and a sudden spike in a word's surprisal (perhaps relative to its unigram surprisal) could be used as a signal of parsing failures. Inviting this possibility, \cite{timkey_eye_2025}, shows in an eyetracking-while-reading study that LM surprisal can account for garden path effects in trials where the disambiguating word does not cause rereading, but drastically underpredicts the rate at which the disambiguating word triggers rereading of previous words. It is possible that difficulty metrics from a limited-beam RNNG could be better predictors of regressive eye movements than standard LMs. While our results suggest that LM surprisal alone cannot fully account for syntactic disambiguation difficulty, then, it remains possible that LMs can provide a full account of processing difficulty in naturalistic and garden path sentences alike, albeit with a more complex linking function that is assumed by surprisal theory.

\section{Conclusion}
This work evaluated whether differences between humans and language models in parse multiplicity---the number of parses that can be considered in parallel---can explain the consistent failure of neural language models to capture the magnitude of garden path effects. Using RNNGs with word synchronous beam search as a next word prediction language model, we showed that limiting the number of simultaneously maintained parses leads to garden path effects that are larger in magnitude, but these memory-limited models still fall far short of matching the full human magnitude, even under extreme conditions that enforce a single, incorrect interpretation. These findings suggest that the gap between model predictions and human behavior cannot be resolved by adjusting parse multiplicity alone, and instead points to a more fundamental limitation of single-stage, surprisal-based accounts. Accurately modeling human sentence processing will likely require incorporating additional mechanisms that go beyond word predictability, such as explicit syntactic reanalysis.

\bibliographystyle{apalike}
\bibliography{references, custom}

\newpage
\appendix

\counterwithin{figure}{section}

\section{The action beam width parameter, $k_a$}
\label{sec:action_beam}
Word-synchronous beam search includes a second hyperparameter, the \textit{action beam width} $k_a$ (often denoted simply as $k$), which limits how many intermediate parsing states are considered as the model builds structure between successive words. While this parameter can have a large effect on the search procedure, our hypothesis concerns constraints on the number of competing parses that are actively maintained as each word is observed, rather than constraints on the step-by-step construction of structure \textit{between} two words. Indeed, it is not theoretically necessary to assume that syntactic structure is built via a sequence of discrete intermediate actions at all. One could instead imagine that the relevant structure between two words is computed in a single step \citep[e.g.][Ch. 8]{berwick1984grammatical, hale_automaton_2014}, in which case constraints on $k_a$ would have no clear cognitive interpretation. Accordingly, we fix $k_a$ to a large value ($k_a = 1000$) in all analyses, so as not to artificially restrict the space of intermediate derivations between two words.

\section{Approximating a fully-parallel parser}
\label{sec:full_parallel_cond}
The \textsc{Full-Parallel} condition serves as a baseline to calculate surprisals from the RNNG using a beam that is guaranteed to contain the globally correct parse before and after disambiguation. We first obtained the output of beam search on the original ambiguous garden path sentences at $k_w = 1000$. We refer to the set of parses obtained by beam search on the original sentences as $b$. We then constructed new versions of the garden path sentences where the first verb in the ambiguous condition is replaced with a verb that eliminates the temporary ambiguity. The globally correct parse of the modified sentences is structurally identical to the globally correct parse of the original ambiguous garden path version:

\noindent \textbf{MV/RR}: The suspect \sout{sent} \textbf{given} the file \textit{deserved} further investigation. \\
\noindent \textbf{NP/S}: The suspect \sout{showed} \textbf{said} the file \textit{deserved} further investigation.\\
\noindent \textbf{NP/Z}: Because the suspect \sout{changed} \textbf{died} the file \textit{deserved} further investigation.\\

We then carry out beam search with $k_w = 1000$ on these modified versions of the sentences, which produces the set of parses $b'$. This step is only done to obtain the parses themselves, but their associated probabilities are ignored. We then manually confirmed for each item that at least one of the 1000 parses was consistent with the globally correct interpretation (i.e. the RR, S, and Z interpretations of the MV/RR, NP/S and NP/Z sentences respectively) at the end of the sentence. For each parse in $b'$, we then replace the unambiguous verb with the ambiguous verb from the original sentence. This gives us a set of parses, $b''$ that contain the \textit{words} of the original ambiguous sentences, but with the set of \textit{structures} we would expect to see if there was never any temporary ambiguity in the first place. The parse probabilities and next word surprisals can then be computed directly from the RNNG for all $b''$. Finally, we compute marginal next-word probabilities by marginalizing over the union of the original parses, $b$, and the counterfactual parses, $b''$ (we ensure that duplicate parses from $b$ and $b''$ are counted only once). This gives us next word surprisals that are guaranteed to include the globally correct interpretation of the sentence on the beam prior to disambiguation. As with the \textsc{Forced Garden Path} condition, we use the $k=1000$ filler models when converting surprisals from this condition into predicted reading times.

\section{Full Results from RNNGs trained on BabyLM corpus} 

\begin{figure*}[tb]
\includegraphics[width=\columnwidth]{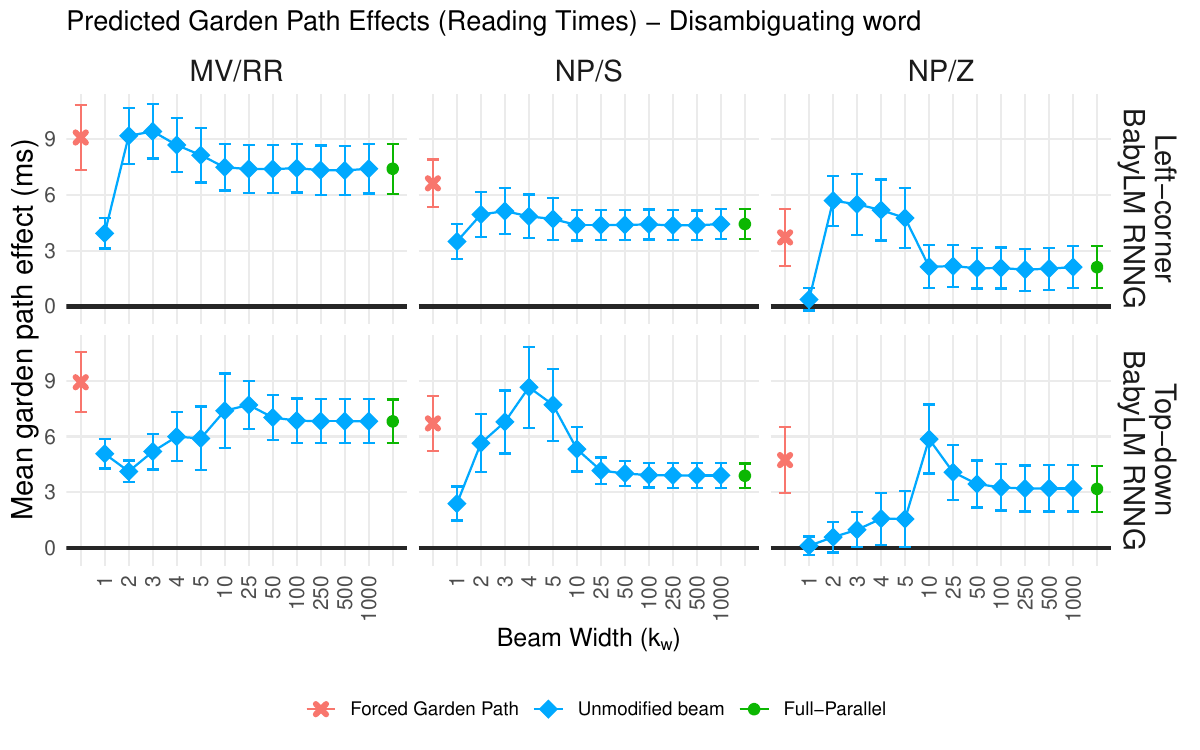} 
\caption{Predicted garden path effects from the BabyLM RNNGs at the critical word, measured in milliseconds of reading time. Empirical effect sizes are represented as horizontal bands. Error bars represent 95\% credible intervals, with adjustments for repeated measures across RNNG model seeds}
\label{fig:gpes_by_beamsize_rts_0_babylm}
\end{figure*}

\begin{figure*}[tb]
\includegraphics[width=\columnwidth]{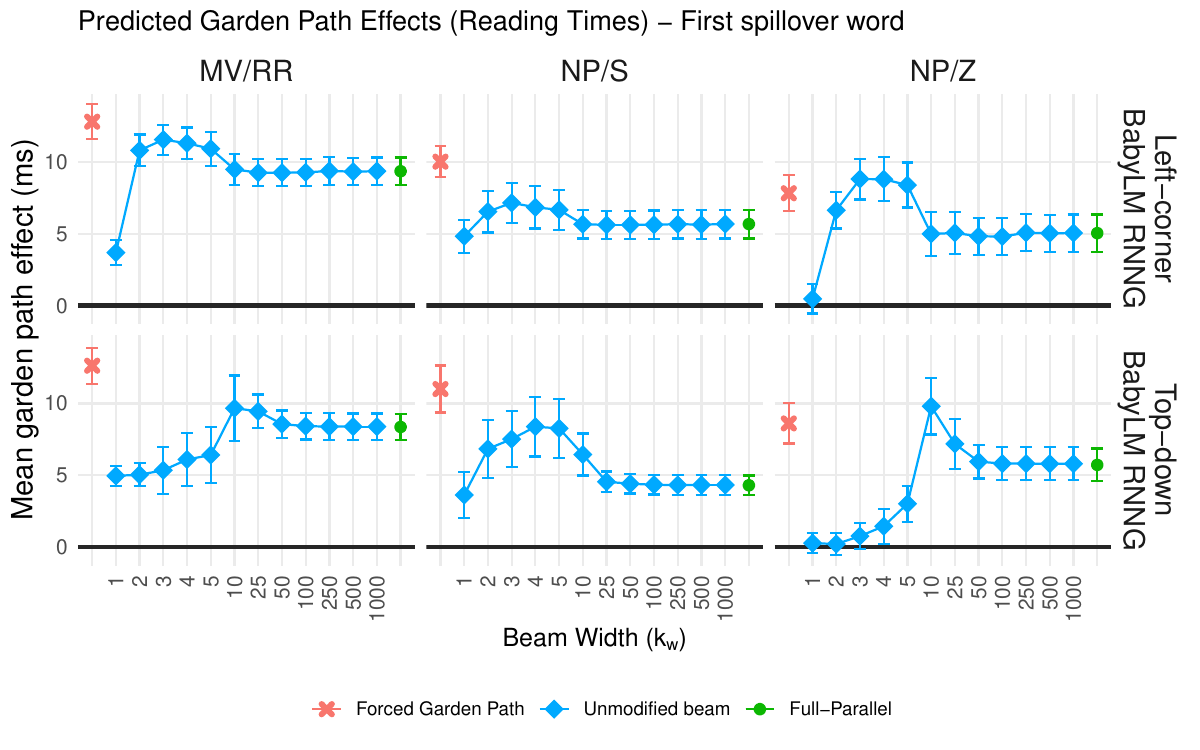} 
\caption{Predicted garden path effects from the BabyLM RNNGs at the first spillover word, measured in milliseconds of reading time. Empirical effect sizes are represented as horizontal bands. Error bars represent 95\% credible intervals, with adjustments for repeated measures across RNNG model seeds}
\label{fig:gpes_by_beamsize_rts_1_babylm}
\end{figure*}

\begin{figure*}[tb]
\includegraphics[width=\columnwidth]{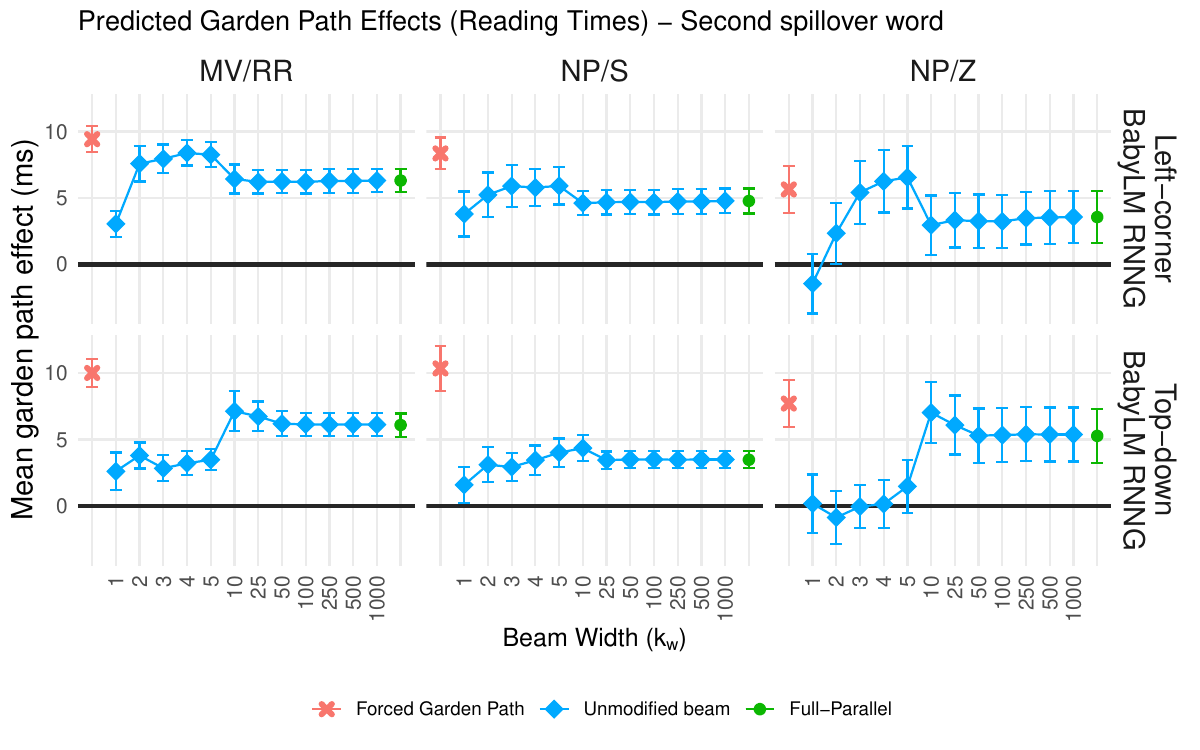} 
\caption{Predicted garden path effects from the BabyLM RNNGs at the second spillover word, measured in milliseconds of reading time. Empirical effect sizes are represented as horizontal bands. Error bars represent 95\% credible intervals, with adjustments for repeated measures across RNNG model seeds}
\label{fig:gpes_by_beamsize_rts_2_babylm}
\end{figure*}

Garden path effects measured as a difference in predicted reading times, are summarized for the disambiguating word in Figure \ref{fig:gpes_by_beamsize_rts_0_babylm}, the first spillover word in Figure \ref{fig:gpes_by_beamsize_rts_1_babylm}, and the second spillover word in Figure \ref{fig:gpes_by_beamsize_rts_2_babylm}.

\section{Results from RNNGs trained on BLLIP corpus}
\begin{figure*}[tb]
\includegraphics[width=\columnwidth]{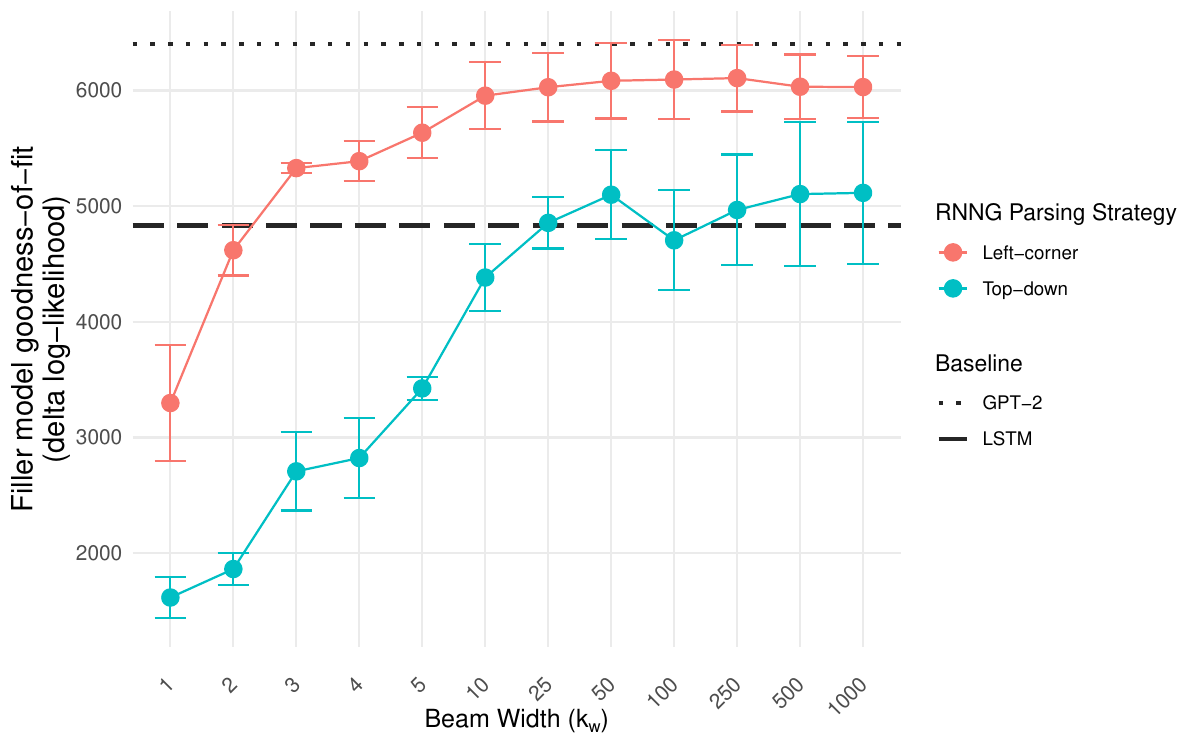} 
\caption{Goodness of fit (Delta log-likelihood over baseline model; higher is better) to filler sentences improves when surprisal is estimated using larger beam widths. Results from the language models investigated in \cite{huang_large-scale_2024}, GPT-2 small \citep{radford2019language} and an LSTM language model \citep{gulordava_colorless_2018} are provided for reference. Error bars are standard errors over each condition's 5 model seeds.}
\label{fig:filler_dll_bllip}
\end{figure*}

In the RNNGs trained on the BLLIP corpus, the goodness-of-fit to reading times in the filler sentences showed a remarkably similar pattern to the models trained on the BabyLM corpus. Both provided a better fit to reading times as beam size increased, with the left-corner models consistently outperforming the top-down models, particularly at smaller beam sizes (Figure~\ref{fig:filler_dll_bllip}).

\begin{figure*}[tb]
\includegraphics[width=\columnwidth]{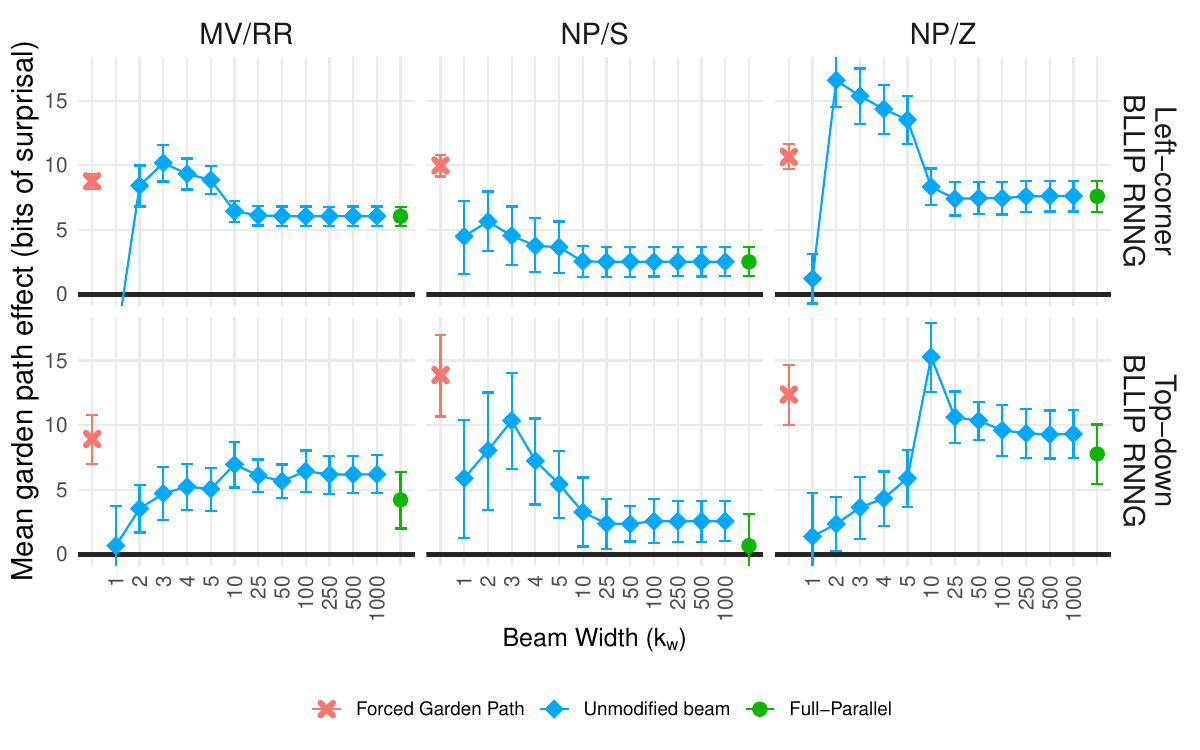} 
\caption{Predicted garden path effects from the RNNGs trained on the BLLIP corpus. Effects are measured in surprisal, summed across the disambiguating word and two spillover words. Error bars represent 95\% credible intervals, estimated from Bayesian regression models, described in Appendix \ref{sec:regression_info}.}
\label{fig:gpes_by_beamsize_surp_bllip}
\end{figure*}

\begin{figure*}[tb]
\includegraphics[width=\columnwidth]{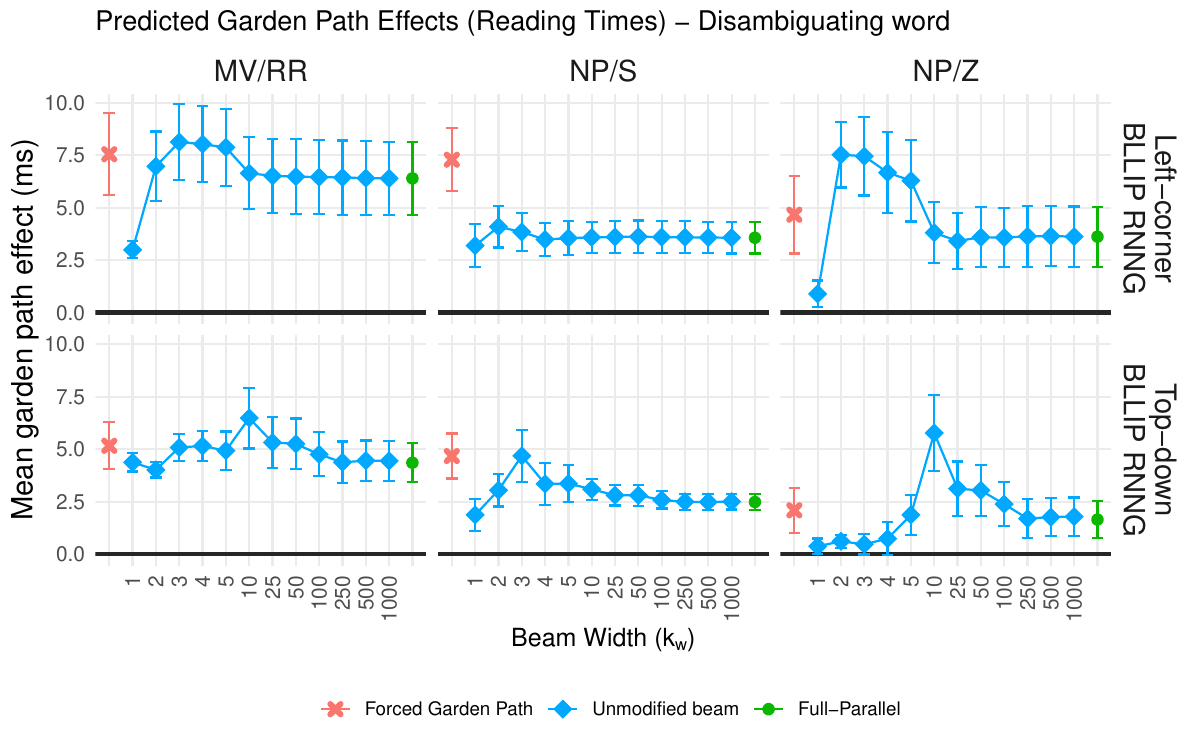} 
\caption{Predicted garden path effects from the BLLIP RNNGs at the critical word, measured in milliseconds of reading time. Empirical effect sizes are represented as horizontal bands. Error bars represent 95\% credible intervals, with adjustments for repeated measures across RNNG model seeds}
\label{fig:gpes_by_beamsize_rts_0_bllip}
\end{figure*}

Surprisals from the BLLIP models exhibited a very similar pattern to those from the BabyLM models. Effects were consistently larger in the \textsc{Forced Garden Path} condition than the \textsc{Full-Parallel} condition, and the beam width that predicted the largest effect magnitude was always between two and ten (Figure \ref{fig:gpes_by_beamsize_surp_bllip}). One possible exception to the general pattern, and a qualitative divergence from the BabyLM models, was the MV/RR construction in the top-down models, where the largest predicted effect at $k_w = 10$ was only marginally larger than the effect at the largest beam sizes. The qualitative pattern of predicted reading times was identical to that of the raw surprisal effects, but predicted effect sizes were once again orders of magnitude smaller than the empirical effects at the disambiguating word (Figure~\ref{fig:gpes_by_beamsize_rts_0_bllip}), the first spillover word (Figure \ref{fig:gpes_by_beamsize_rts_0_bllip}), and the second spillover word  (Figure \ref{fig:gpes_by_beamsize_rts_0_bllip}).

\label{sec:bllip_results}

\begin{figure*}[tb]
\includegraphics[width=\columnwidth]{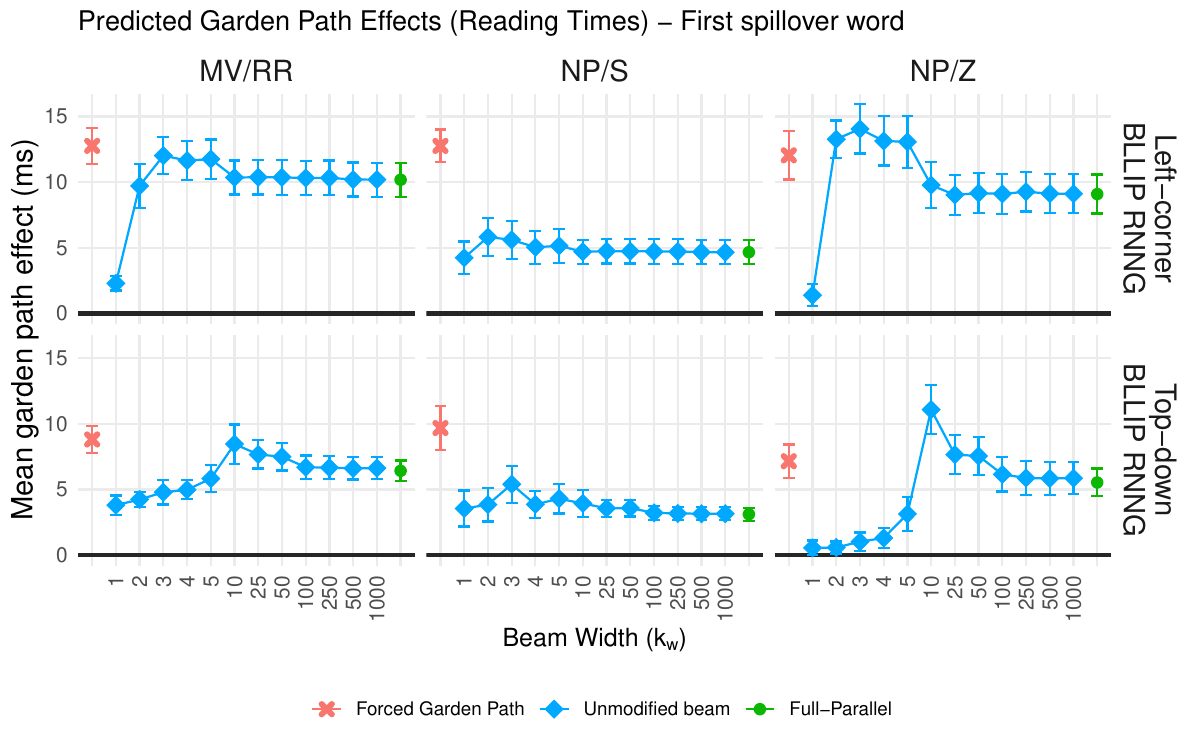} 
\caption{Predicted garden path effects from the BLLIP RNNGs at the first spillover word, measured in milliseconds of reading time. Empirical effect sizes are represented as horizontal bands. Error bars represent 95\% credible intervals, with adjustments for repeated measures across RNNG model seeds}
\label{fig:gpes_by_beamsize_rts_1_bllip}
\end{figure*}

\begin{figure*}[tb]
\includegraphics[width=\columnwidth]{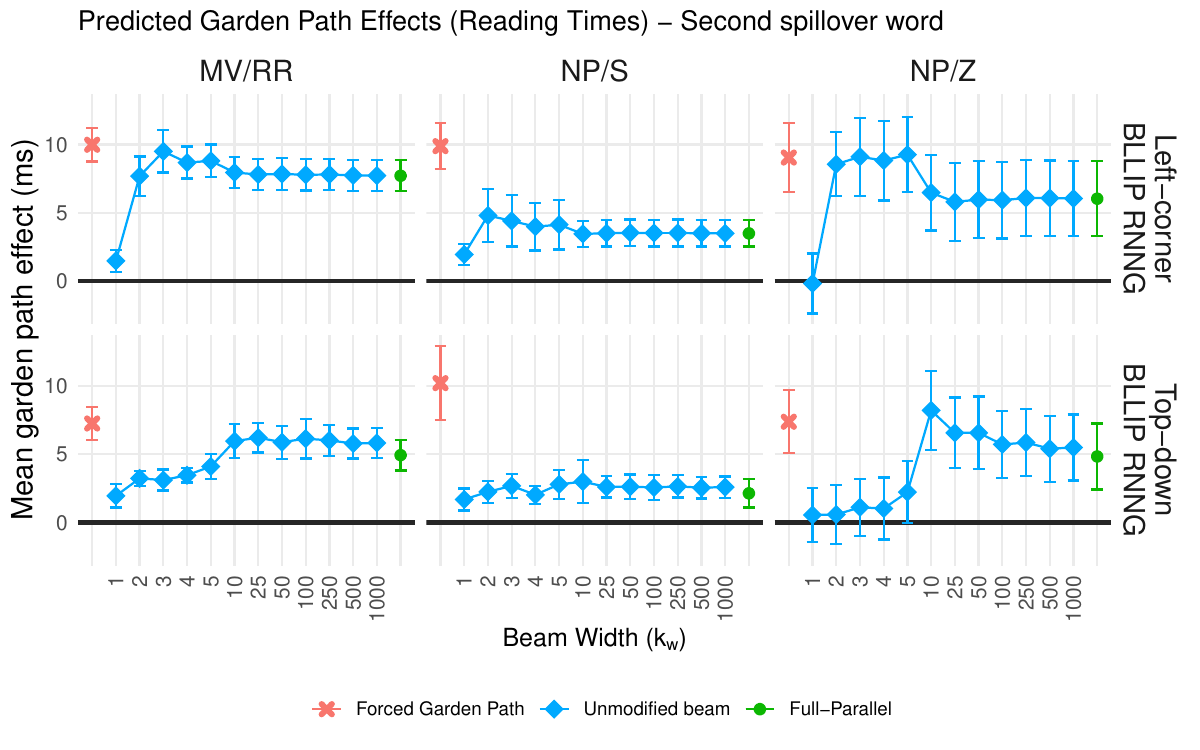} 
\caption{Predicted garden path effects from the BLLIP RNNGs at the second spillover word, measured in milliseconds of reading time. Empirical effect sizes are represented as horizontal bands. Error bars represent 95\% credible intervals, with adjustments for repeated measures across RNNG model seeds}
\label{fig:gpes_by_beamsize_rts_2_bllip}
\end{figure*}

\section{RNNG Parsing performance on Penn Treebank}
\label{sec:parsing_performance}
For each of our 20 RNNGs, we obtain the most probable parse of each sentence in the Penn Treebank test set (partition 23) using word synchronous beam search with $k_a = 2000$ and $k_w = 200$, and compute the labeled F1 score against the gold parse, averaging this score across 5 random seeds from each combination of training corpus and parsing strategy.

The left-corner and top-down models trained on the BabyLM corpus obtain average F1 scores of \textbf{0.923} and \textbf{0.890} respectively. The left-corner and top-down and models trained on the BLLIP corpus obtain average F1 scores of \textbf{0.887} and \textbf{0.850} respectively. These scores are surprisingly competitive with state-of-the-art constituency parsers (e.g. \citet{kitaev_constituency_2018}, which we used to parse our training corpora, obtain an F1 score of 0.95). Unlike our RNNG models, these state-of-the-art parsers are typically trained directly on in-distribution, gold-annotated parses from the training partition of the Penn Treebank. Overall, these results give us reasonable confidence that our trained models can provide high quality estimates of parse probabilities.

\section{Statistical modeling}
\label{sec:regression_info}
\subsection{Garden path effect estimation}
\subsubsection{Empirical garden path effects}
Empirical garden path effects are estimated from reading time data using Bayesian mixed effects linear regression from the \texttt{brms} package in R. We do not fit new empirical models for our analyses; we use those provided by \cite{huang_large-scale_2024}. These models were fit for each region of interest: The disambiguating word, and each of the two immediately following words. Each model includes all three constructions. The three constructions were coded using treatment coding, where the first variable \texttt{NPS} contrasted the NP/S construction from the others (MV/RR:~$0$, NP/Z:~$0$, NP/S:~$1$), and the second variable \texttt{NPZ} contrasted the NP/Z construction from the others (MV/RR:~$0$, NP/S:~$0$, NP/Z:~$1$). The regression models included the main effects of ambiguity, \texttt{NPS}, and \texttt{NPZ}, as well as the two-way interactions \texttt{Ambiguity:NPS} and \texttt{Ambiguity:NPZ}. The models also included by-participant and by-item random slopes for all main and interaction effects, as well as by-participant and by-item random intercepts:

\begin{singlespace}
\begin{align*}
    Measure\text{ }\sim \text{ } & ambiguity \bdot (NPS \mplus NPZ)\mplus \\
     & \bigl(1 \mplus ambiguity \bdot (NPS \mplus NPZ) \parallel subject\bigr)\mplus\\
     & \bigl(1 \mplus ambiguity \bdot(NPS \mplus NPZ) \parallel item\bigr)
\end{align*}
\end{singlespace}

Per-construction ambiguity effect estimates and 95\% credible intervals were estimated for each construction using the posterior samples from the trained \texttt{brms} model.

\subsection{Filler models}
Our method for fitting filler models was identical to that described in \cite{huang_large-scale_2024}. All models are fit with linear mixed-effects regression models from the \texttt{lme4} package in R to reading time data from the filler sentences in the SAP Benchmark self-paced reading dataset \citep{huang_large-scale_2024}. We first fit a baseline filler model that predicts reading times on the filler sentences from a set of baseline factors including word position, word length, word frequency, and the interaction of frequency and length. The model also includes random intercepts for participants and items. All predictors are scaled to mean zero and standard deviation of one. The baseline filler model regression formula is described below:

\begin{singlespace}
\begin{align*}
    RT(w_i) \text{ }\sim\text{ } & Logfreq(w_i)\bdot Length(w_i)\mplus \\ & Logfreq(w_{i-1}) \bdot Length(w_{i-1})\mplus \\ &
    Logfreq(w_{i-2}) \bdot Length(w_{i-2})\mplus \\
     & Position(w_i)\mplus \\
     & (1 \mid particpant)\mplus \\
     & (1 \mid item)
\end{align*}
\end{singlespace}

\noindent The full filler models additionally include surprisals from the current word ($w_i$) and the previous two words, as well as random surprisal slopes for participants. One filler model is fit for each combination of RNNG seed, parsing strategy, training corpus, and beam width:

\begin{singlespace}
\begin{align*}
    RT(w_i) \text{ }\sim\text{ } & Surp(w_i) \mplus Surp(w_{i-1})\mplus Surp(w_{i-2})\mplus \\
     & Logfreq(w_i)\bdot Length(w_i)\mplus \\
     & Logfreq(w_{i-1}) \bdot Length(w_{i-1})\mplus \\
     &Logfreq(w_{i-2}) \bdot Length(w_{i-2})\mplus \\
     & Position(w_i)\mplus \\
     & (1 + Surp(w_i) \mplus Surp(w_{i-1}) + Surp(w_i) \mplus Surp(w_{i-2}) \mid particpant)\mplus \\
     & (1 \mid item)
\end{align*}
\end{singlespace}

\noindent Each full filler model includes surprisals from a single RNNG model at a single beam width. Goodness of fit to the training data calculated by subtracting the log likelihood of the filler data given the baseline model from the log likelihood of the filler data given a full filler model.

\subsubsection{Surprisal-predicted garden path effects}
When estimating garden path effects in raw surprisal, we fit a single Bayesian regression model for each unique combination of beam width and parsing strategy. This model predicts the summed surprisal of the disambiguating word, first spillover and second spillover words from the same ambiguity and construction predictors as in the empirical model. The random effects structure for items is the same as in the empirical model, and we also include a random intercept for model seed:

\begin{singlespace}
\begin{align*}
    Surprisal\text{ }\sim \text{ } & ambiguity \bdot (NPS \mplus NPZ)\mplus \\
     & \bigl(1  \mid seed\bigr)\mplus\\
     & \bigl(1 \mplus ambiguity \bdot(NPS \mplus NPZ) \parallel item\bigr)
\end{align*}
\end{singlespace}

When estimating garden path effects as differences in predicted reading times, we follow \cite{huang_large-scale_2024} in first using the trained filler models to predict reading times in the garden path stimuli from surprisals. For the unambiguous stimuli in the \textsc{Forced Garden Path} and \textsc{Full-Parallel} conditions, we simply use surprisals estimated at $k_w = 1000$, rather than manually pruning the beam to have only the ultimately correct interpretation (as in practice, nearly all of the parses on the beam in the unambiguous sentences are consistent with the globally correct interpretation before and after the disambiguating word, see Figure~\ref{fig:interp_probs_topdown_babylm_unambig} of Appendix~\ref{sec:interp_probs}). We then estimate garden path effect sizes from predicted reading times using Bayesian regression models with the same formula as the empirical models, with an additional random intercept for model seeds. We fit one model for each unique combination of training corpus, parsing strategy and beam size, with the following formula:

\begin{singlespace}
\begin{align*}
    RT(w_i)\text{ }\sim \text{ } & ambiguity \bdot (NPS \mplus NPZ)\mplus \\
     & \bigl(1 \mplus ambiguity \bdot (NPS \mplus NPZ) \parallel subject\bigr)\mplus\\
     & \bigl(1 \mplus ambiguity \bdot(NPS \mplus NPZ) \parallel item\bigr)\mplus\\
     & \bigl(1  \mid seed\bigr)\\
\end{align*}
\end{singlespace}

95\% credible intervals are estimated using the same methodology as with the empirical effect estimates.

\section{Interpretation probabilities by beam width}
\label{sec:interp_probs}
\begin{figure*}[tb]
\includegraphics[width=0.95\columnwidth]{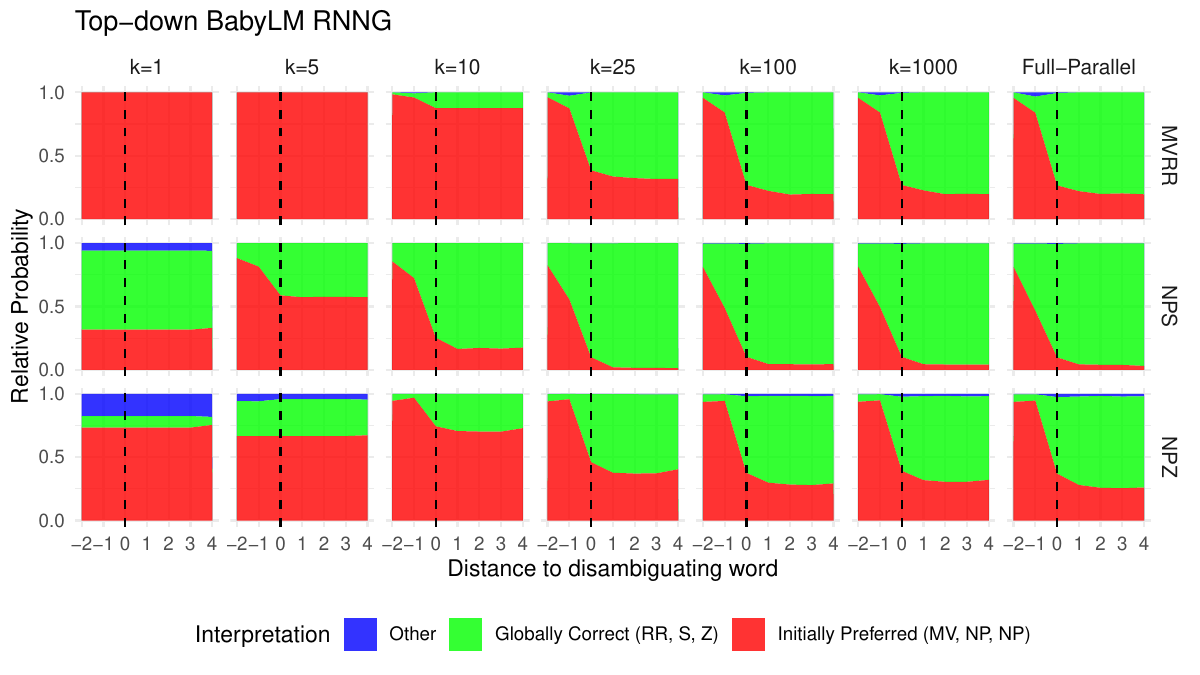} 
\caption{Relative probabilities of each interpretation in the ambiguous condition from the top-down RNNGs trained on the BabyLM dataset, split by beam width ($k_w$), and construction. Within each plot, the horizontal bar denotes the disambiguating word, and the x-axis is labeled based on the word's index aligned to the disambiguating word (i.e. ``1'' corresponds to the first spillover word, and ``-1'' corresponds to the word immediately preceding the disambiguating word).}
\label{fig:interp_probs_topdown_babylm_ambig}
\end{figure*}

\begin{figure*}[tb]
\includegraphics[width=0.95\columnwidth]{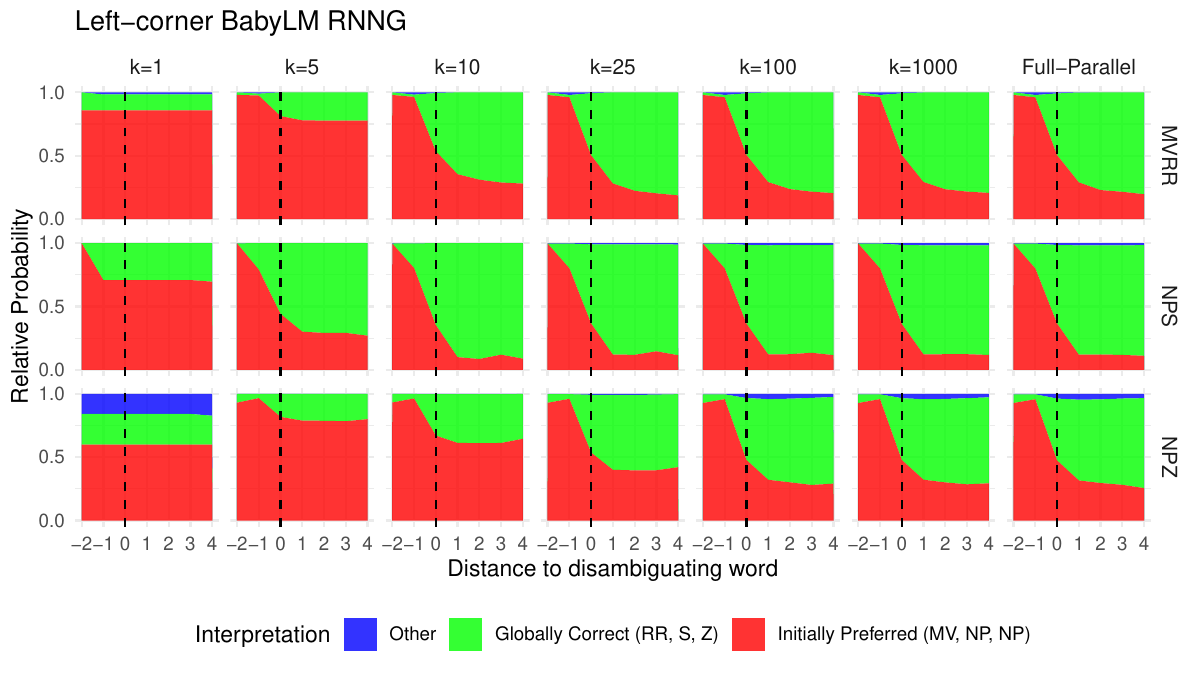} 
\caption{Relative probabilities of each interpretation in the ambiguous condition from the left-corner RNNGs trained on the BabyLM dataset, split by beam width ($k_w$), and construction. Within each plot, the horizontal bar denotes the disambiguating word, and the x-axis is labeled based on the word's index aligned to the disambiguating word (i.e. ``1'' corresponds to the first spillover word, and ``-1'' corresponds to the word immediately preceding the disambiguating word).}
\label{fig:interp_probs_leftcorner_babylm_ambig}
\end{figure*}

\begin{figure*}[tb]
\includegraphics[width=0.95\columnwidth]{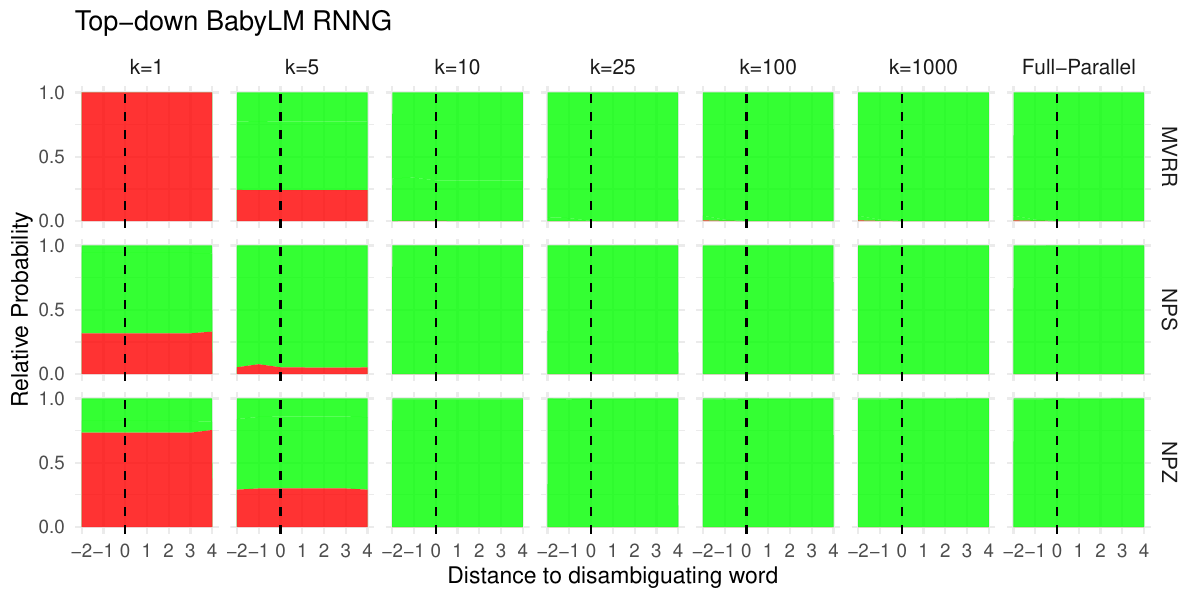} 
\caption{Relative probabilities of each interpretation in the unambiguous condition from the top-down RNNGs trained on the BabyLM dataset, split by beam width ($k_w$), and construction. Within each plot, the horizontal bar denotes the disambiguating word, and the x-axis is labeled based on the word's index aligned to the disambiguating word (i.e. ``1'' corresponds to the first spillover word, and ``-1'' corresponds to the word immediately preceding the disambiguating word).}
\label{fig:interp_probs_topdown_babylm_unambig}
\end{figure*}

\begin{figure*}[tb]
\includegraphics[width=0.95\columnwidth]{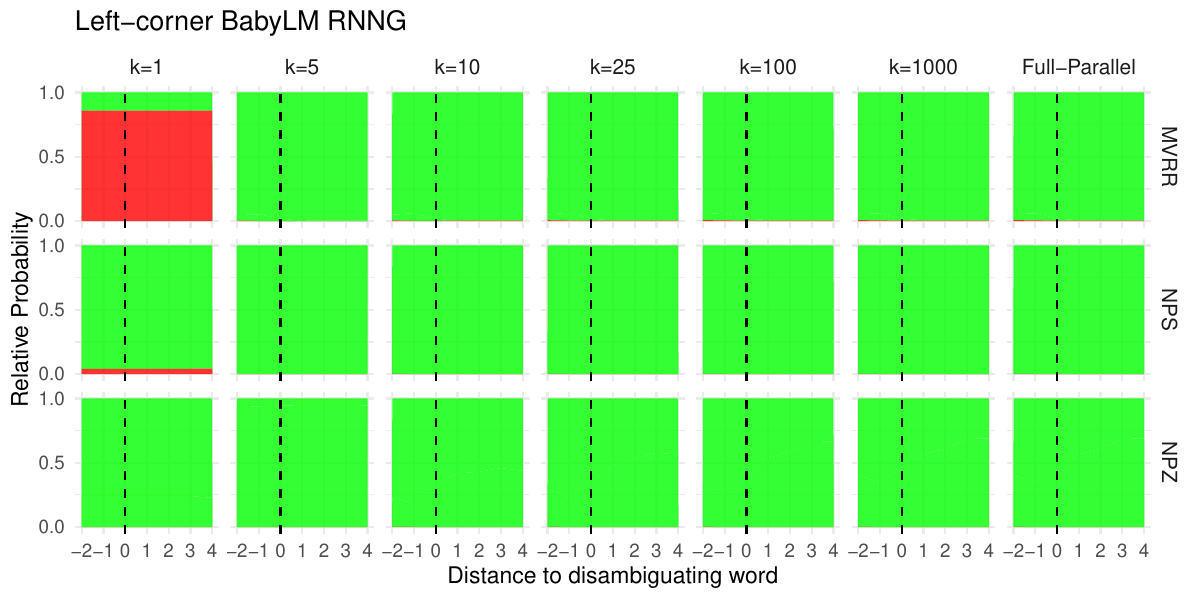}
\caption{Relative probabilities of each interpretation in the unambiguous condition from the left-corner RNNGs trained on the BabyLM dataset, split by beam width ($k_w$), and construction. Within each plot, the horizontal bar denotes the disambiguating word, and the x-axis is labeled based on the word's index aligned to the disambiguating word (i.e. ``1'' corresponds to the first spillover word, and ``-1'' corresponds to the word immediately preceding the disambiguating word).}
\label{fig:interp_probs_leftcorner_babylm_unambig}
\end{figure*}

In this analysis, we manually inspect the word-by-word probabilities assigned by the RNNGs to each interpretation of the garden path sentences (the initially preferred or ultimately correct interpretation) as a function of the word beam width parameter $k_w$. As each word is observed, the word synchronous beam search algorithm outputs a set of the $k_w$ highest probability parses of those words. We manually bin the parses at each word by whether they are consistent with the initially preferred interpretation (the NP interpretation in the NP/Z and NP/S constructions and the MV interpretation in the MV/RR sentences), or the globally correct interpretation. We sum the probability of all the parses in each bin, and normalize by the total probability of all parses. These probabilities are averaged across RNNG seeds and items. Results from ambiguous sentences are provided in Figure~\ref{fig:interp_probs_topdown_babylm_ambig} for the top-down BabyLM models, and Figure~\ref{fig:interp_probs_leftcorner_babylm_ambig} for the left-corner BabyLM models. Results from unambiguous sentences are provided in Figure~\ref{fig:interp_probs_topdown_babylm_unambig} for the top-down BabyLM models, and Figure~\ref{fig:interp_probs_leftcorner_babylm_unambig} for the left-corner BabyLM models.

\newpage

\end{document}